# Identification of structural features in chemicals associated with cancer drug response: A systematic data-driven analysis

Suleiman A Khan[1,*], Seppo Virtanen[1], Olli P Kallioniemi[2], Krister Wennerberg[2], Antti Poso[2,3] and Samuel Kaski[1,4,*]

[1]Helsinki Institute for Information Technology HIIT, Department of Information and Computer Science, Aalto University, 00076 Espoo, Finland. [2]Institute for Molecular Medicine Finland FIMM, University of Helsinki, 00014 Helsinki, Finland. [3]School of Pharmacy, Faculty of Health Sciences, University of Eastern Finland, 70211 Kuopio, Finland. [4]Helsinki Institute for Information Technology HIIT, Department of Computer Science, University of Helsinki, 00014 Helsinki, Finland.

**ABSTRACT**

**Motivation:** Analysis of relationships of drug structure to biological response is key to understanding off-target and unexpected drug effects, and for developing hypotheses on how to tailor drug therapies. New methods are required for integrated analyses of a large number of chemical features of drugs against the corresponding genome-wide responses of multiple cell models.

**Results**: In this paper, we present the first comprehensive multi-set analysis on how the chemical structure of drugs impacts on genome-wide gene expression across several cancer cell lines (CMap database). The task is formulated as searching for drug response components across multiple cancers to reveal shared effects of drugs and the chemical features that may be responsible. The components can be computed with an extension of a very recent approach called Group Factor Analysis (GFA). We identify 11 components that link the structural descriptors of drugs with specific gene expression responses observed in the three cell lines, and identify structural groups that may be responsible for the responses. Our method quantitatively outperforms the limited earlier studies on CMap and identifies both the previously reported associations and several interesting novel findings, by taking into account multiple cell lines and advanced 3D structural descriptors. The novel observations include: previously unknown similarities in the effects induced by 15-delta prostaglandin J2 and HSP90 inhibitors, which are linked to the 3D descriptors of the drugs; and the induction by simvastatin of leukemia-specific anti-inflammatory response, resembling the effects of corticosteroids.

**Contact:** suleiman.khan@aalto.fi, samuel.kaski@aalto.fi

## 1 INTRODUCTION

Modeling and understanding the diverse spectrum of cellular responses to drugs is one of the biggest challenges in chemical systems biology. Some of the responses can be predicted for targeted drugs, which have been designed to bind to a specific protein that triggers the biological response. The binding of a drug to a target largely depends on the structural correspondence of the drug molecule and the binding cavity of the target molecule which can be modeled in principle, given ample computational resources. Off-target effects are harder to predict. They are dependent on the cell types, individual genetic characteristics and cellular states making the spectrum of responses overwhelmingly diverse. The less well known the drug's mechanism of action and the characteristics of the disease, the harder the prediction from first principles becomes. The most feasible way to approach this challenge in an unbiased way, which does not require prior knowledge of all on- and off-target interactions of drugs, is to collect systematic measurements across different drugs, cell types, and diseases, and search for response patterns correlating with the characteristics of the drugs. The patterns found can be used as evidence for hypotheses on underlying action mechanisms, or directly in predicting the responses.

The Connectivity Map (CMap; Lamb *et al.*, 2006) described the basis for a data-driven study of drug-effect relationships at a genome-wide level. CMap hosts the largest collection of high-dimensional gene expression profiles derived from treatment of three different human cancer cell lines with over one thousand drugs. The CMap data have been used in a multitude of studies revealing new biological links between drugs and between drugs and diseases. Genome-wide gene expression responses from the CMap have been used to discover clusters of drugs having similar mechanisms of action, resulting in novel findings, such as effects of heat shock protein (HSP) inhibitors and identification of modulators of autophagy (Iorio *et al.*, 2010). The CMap data have also been successfully used in large scale integrative studies including the analysis of regulation of drug targets (Iskar *et al.*, 2010), hERG annotations to predict novel inhibitors (Babcock *et al.*, 2013) and drugs' interactions with protein networks (Laenen *et al.*, 2013).

Quantitative structure activity relationship analysis (QSAR; Cramer *et al.*, 1988) is a widely adopted approach to studying drug responses. Traditionally, univariate biological activities are predicted using a range of methods, including classical regression, Support Vector Machines, and Random Forests. The key challenge when moving from traditional QSAR to systems wide analysis of chemical effects is how to relate structural features to genome-wide cellular responses.

Integration of chemical structures with genome-wide responses has become a major research direction in Chemical Systems Biology (Iskar *et al.*, 2012; Xie *et al.*, 2012). Keiser *et al.* (2009) studied structural similarities between ligand sets while Klabunde *et al.* (2005) used protein-ligand complexes to predict off-targets. To infer potential indications for drugs, Gottlieb *et al.* (2011) combined similarities from chemical structures, gene expression profiles, protein targets and several other datasets. Atias *et al.* (2011) modeled linkage between structural descriptors of drugs and their side effects using Canonical Correlation Analysis (CCA; Hotelling, 1936). Structures have also been used with genomic datasets to predict toxicity and complex adverse drug reactions (Russom *et*

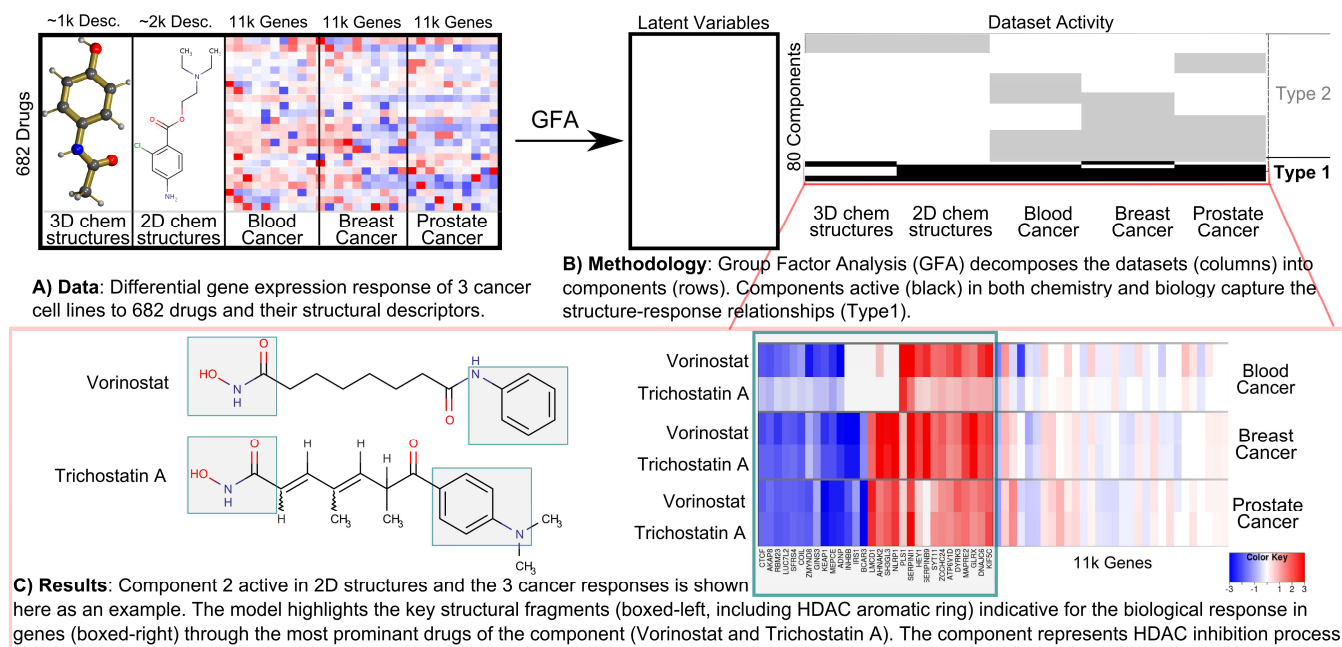

**Fig. 1.** Overview of the symmetric multi-structure to multi-response decomposition. **A**) The five datasets spanning 682 paired drugs are **B**) decomposed into components by GFA. Components of Type 1 represent shared patterns in both chemistry and biology while Type 2 describes biology-only or chemistry-only variation (not as useful in our case). **C**) Each shared component identifies key structures and genes of an underlying biological process.

*al.*, 2013). Recently, Menden *et al.* (2013) combined structures of drugs and mutation information of cell lines to predict drug cytotoxicity in a series of cell lines.

Relationships between structural descriptors of drugs and their gene expression profiles have also been studied. Cheng *et al.* (2010) examined similarities between chemical structures and molecular targets of 37 drugs that were clustered based on their bioactivity profiles. Low *et al.* (2011) classified 127 rat liver samples to toxic vs non-toxic responses, based on combined drug-induced expression profiles and chemical descriptors, and identified chemical substructures and genes that were responsible for liver toxicity. In a broader setting, when the goal is to find dependencies between two data sources (chemical structures and genomic responses), correlation-type approaches match the goal directly, and have the additional advantage that a pre-defined classification is not required. Khan *et al.* (2012) generalized structure-response analysis to multivariate correlations with Canonical Correlation Analysis (CCA) on the CMap. Due to the limitations of classical CCA, their study was restricted to a limited set of descriptors (76) and genomic summaries (1321 genesets), and did not attempt to take into account the data from three separate cell lines.

In this paper we present the first probabilistic approach to the problem of integrated analysis of effects of chemical structures across genome-wide responses in multiple model systems. We extend the earlier work in three major ways: (i) Instead of using only two data sources (as in classical CCA), we used the recent Bayesian Group Factor Analysis method (GFA; Virtanen *et al.*, 2012) that generalizes the analysis to multiple sources, here three cell lines and two sets of chemical descriptors. (ii) Our Bayesian treatment with feature-level priors enabled us to cope better with the uncertainties in the high-dimensional data. (iii) We included a more informative set of 3D chemical descriptors to complement the widely used 2D fingerprints which are recognized to only explain limited aspects of drugs (Schneider, 2010).

Our goal was to uncover the big picture of relationships between chemical structure parameters and genome-wide responses, in a data-driven fashion (Fig. 1). The data came from CMap, 11,327 gene-wide responses in three cell lines (HL60-Blood Cancer/Leukemia, MCF7-Breast Cancer and PC3-Prostate Cancer; Lamb *et al.*, 2006), and from two sets of chemical descriptors: 780 3D Pentacle descriptors of drugs (Duran *et al.*, 2008) and 2,769 functionally relevant structural fragments (FCFP4; Glen *et al.*, 2006) as 2D fingerprints of the drugs. These five datasets consist of samples from the 682 drug treatments, coupled by the detailed drug identity. We analyzed the statistical relationships between the datasets by decomposing them into a set of interpretable components. Our method quantitatively outperformed previous studies, thereby validating the approach. We rediscovered findings reported earlier as well as identified novel drug associations and detailed structure-response relationships.

## 2 METHODS

### 2.1 Gene Expression Datasets

We used the Connectivity Map (Lamb *et al.*, 2006) gene expression data as a measure of the biological response of the three cancer cell lines to drug treatments, forming the gene expression datasets. The Connectivity Map hosts over 7100 gene expression profiles including technical replicates treated with 1309 drugs and is the largest available resource of its kind. Responses from a subset of these drugs (682) were measured on all of the three cell lines, namely HL60 (Leukemia), MCF7 (Breast Cancer) and PC3 (Prostate Cancer cell line).

We obtained the raw gene expression profiles from the Connectivity Map and used the data from the most abundant microarray platform (HT-HG-U133A). The data were preprocessed using RMA (Irizarry *et al.*, 2003) and *drug-treatment vs. control* (log2) differential expression was calculated batch-wise (Khan *et al.*, 2012). Technical replicates were merged by taking the mean of each gene. This resulted in gene expression profiles for the 682 drugs having measurements over all three cell lines. To reduce noise we approximated the approach of Iorio *et al.* (2010) for our setting, by retaining the expression of top 2000 up and down regulated genes for each sample, while considering the rest to be noise (set to zero). These profiles formed three biological response datasets (one for each cell line), each being a differential gene expression matrix of 682 drugs times 11,327 genes.

## 2.2 Chemical Descriptor Datasets

The chemical space of drugs was represented using two different types of chemical descriptors, namely the 2D fingerprints *'FCFP4'* and 3D descriptors *'Pentacle'*. The FCFP4 (Functional Connectivity Fingerprints of radius 4; Glen *et al.*, 2006) are circular topological fingerprints designed specifically for structure-activity modeling and similarity searching. They are rapidly computable and heavily used in a wide variety of applications (Rogers and Hahn, 2010). Each dimension of the fingerprints represents a certain *2-dimensional* fragment of the compounds, interpretable as presence of certain substructures, typically stereochemical information, and allows easy visual inspection of structures. Therefore, FCFP4 can be used to identify the core 2D substructures that make compounds structurally similar and are responsible for biological activity.

The more complex 3-dimensional descriptors Pentacle (Duran *et al.*, 2008) capture the functional properties of the compounds using molecular interaction fields. They are able to group together compounds with very dissimilar chemical structures and yet having the same type of molecular field properties. This is especially important in our study where the aim is to find small molecules that share biological functions despite structural dissimilarity. Most of the traditional fingerprints, like MACCS and FCFP4, are superior to recognize 2D structural similarity but unfortunately unable to recognize structurally unrelated and yet biologically similar compounds binding into the same binding pocket. The opposite is true with most (if not all) field-based similarity methods like Pentacle, which find more effective distant similarities; therefore we decided to combine both approaches. In the earlier work, Khan *et al.* (2012) had used VolSurf descriptors to represent molecular properties. While VolSurf is an optimal method for physicochemical properties estimation it is not able to describe pharmacophore features extensively, unlike the Pentacle descriptors, and thus is not an option in our study.

**Pentacle** field distance descriptors were computed using Pentacle v 1.0.4 (http://www.moldiscovery.com/soft_pentacle.php), by Molecular Discovery. The descriptors were calculated for all the available 10 probe sets, namely D², O², N², T², DO, DN, DT, ON, OT, NT, where D is Dry Probe to represent hydrophobic interactions, O is carbonyl oxygen probe to represent H-bond donor feature of the molecules and N flat probe of Nitrogen is the H-bond acceptor while T is TIP probe representing shape of the molecule, in terms of steric hot spots. For each probe set 78 descriptors were obtained, representing the interaction potentials of probes at different distances, resulting in 780 descriptors in total. Higher interaction potential values indicate stronger interaction of compounds with Pentacle probes. This results in a 682 x 780 data matrix, with each row being a drug and the 780 columns representing the Pentacle descriptors. This forms the first chemical dataset in our study.

The 2D functional connectivity fingerprints **(FCFP4)** represent the chemicals as structural fragments. In FCFP, the fragments are not predefined, rather computed dynamically and thus can represent variation in novel structures. The FCFP4 fingerprints were computed using Pipeline Pilot Student Edition software (http://accelrys.com/products/pipeline-pilot/), by Accelrys. A total of 2,769 unique structural fragments are found and the fingerprints are represented as a matrix of 682 compounds x 2,769 fragment descriptors. This forms the second chemical dataset in our study.

## 2.3 Model: Group Factor Analysis (GFA)

We search for relationships between chemical descriptors and biological responses, as clues to the key underlying biological processes. Group Factor Analysis (GFA) is a model designed to capture such relationships (statistical dependencies) by explaining a collection of datasets ("views") by a set of factors or components, which form a combined low-dimensional representation (Virtanen *et al.*, 2012). In the *multi-view* setting, each component is active in a subset of the datasets, and is a simplified model of an underlying process visible in those sets. The task solved by GFA is to separate the shared components that capture the structure-biology relationships from the rest of the data: the former are visible in all or a subset of the datasets, whereas components active in a single view describe variation specific to that particular view or noise.

Given a collection of M datasets $X^{(1)} \in R^{N \times D_1} \ldots X^{(m)} \in R^{N \times D_M}$, consisting of N co-occurring samples $x_n^{(m)}$, GFA finds a set of latent components (with upper limit K, see below). Each dataset is assumed to have been generated as a linear combination of latent components $Z \in R^{N \times K}$, with weights of the combination given by a loadings matrix $W^{(m)} \in R^{D_m \times K}$: Assuming normal distributions for simplicity, the model is:

$$\begin{aligned} x_n^{(m)} &\sim N\left(W^{(m)} z_n, \Sigma^{(m)}\right), \\ z_n &\sim N(0, I), \end{aligned} \quad (1)$$

where $z_n$ is the $n^{th}$ row of **Z**, and $\Sigma^{(m)}$ is a diagonal noise covariance matrix. GFA is special in that the projections **W** are required to be group-wise sparse, *i.e.*, all the elements $W_{:,k}^{(m)}$ are set to zero for the components $k$ that are not active in the $m^{th}$ dataset. The components with non-zero projections between two or more views capture dependencies between the views.

To increase the interpretability of the model we *extend* GFA by introducing *element-wise* sparsity in addition to the *group-sparsity* for the projection matrices, matching the biological prior assumption that each process typically activates only a subset of genes. We introduce element-wise Automatic Relevance Determination (ARD; Neal, 1996) prior for the projection weight matrices, pushing irrelevant weight values $W_{d,k}^{(m)}$ towards zero and making each component element-wise sparse. For the group sparsity we apply the group spike and slab prior (Klami *et al.*, 2013) where the binary variable $H_k^{(m)}$ controls the activity of the $k^{th}$ component in the group $m$. The prior is

$$\begin{aligned} W_{d,k}^{(m)} &\sim H_k^{(m)} N\left(0, (\alpha_{d,k}^{(m)})^{-1}\right) + \left(1 - H_k^{(m)}\right) \delta_0, \\ H_k^{(m)} &\sim Bernoulli(\pi_k), \\ \pi_k &\sim Beta(a^\pi, b^\pi), \\ \alpha_{d,k}^{(m)} &\sim Gamma(a^\alpha, b^\alpha). \end{aligned} \quad (2)$$

If $H_k^{(m)}$ becomes zero, all values in $W_{:,k}^{(m)}$ will be set to zero. To complete the model description, we set an uninformative prior to the diagonal elements of the precision matrix $(\Sigma^{(m)})^{-1}$.

We represent our (M=5) datasets as matrices of drugs vs. features. The rows represent the samples (drugs) and the columns are the features (genes or chemical descriptors). Drugs are paired in all the views, *i.e.*, a row in all matrices correspond to the same drug. A total of N = 682 drugs were used in the study. The features of the chemical descriptors, Pentacle (m=1) and FCFP4 (m=2) are $D_1$ = 780 Pentacle probe fields and $D_2$ = 2,769 fragment structures, respectively. The biological responses of the three cell lines (m=3,4,5) are represented by differential expression of $D_m$ = 11,327 genes each.

The hyperparameters of the gamma distribution were set to values $(1e^{-3})$ corresponding to an uninformative symmetric prior, while the hyperparameters of the spike and slab were set to $a^\pi, b^\pi = 1$. All remaining model parameters are learned from the data using Gibbs sampling. The number of components is optimally learned from data by initializing K to be large enough, such that sparsity assumptions push some to be inactive. Here for computational reasons we set K=80, a value significantly larger than the actual number of shared components, and let the noise model represent the rest of the data. For sampling, we ran 10 chains and selected for further analysis the one having its likelihood closest to the mean of non-outlier chains. The first 5000 samples were discarded as the burn-in, and the chain was run for 1000 more iterations, with a thinning factor of 5. The mean value of the samples was used as a representation of the model. As a sanity check, we verified that our shared components had over 70% similarity with the second (non-used) chain.

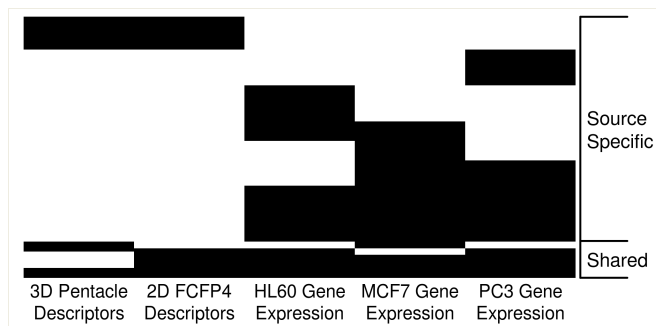

**Fig. 2.** Summary of the GFA components. The plot demonstrates activity (black is active) of each component (y-axis) over the 5 input datasets (x-axis). Each component is active in some or all of the datasets. Components shared (active) by both chemical descriptor and expression datasets capture structure-response relationships.

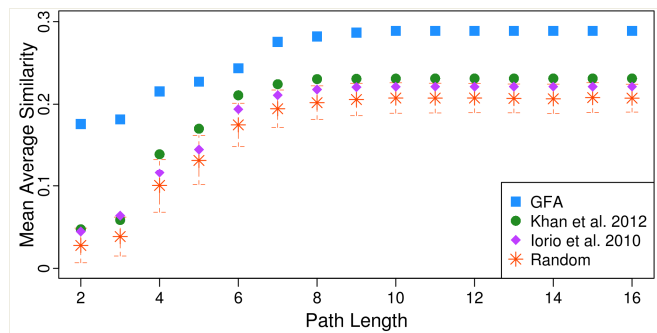

**Fig. 3.** Quantitative Validation of chemical biology similarity of drugs in shared GFA Components. Drugs in the same GFA Component had a consistently higher mean average similarity (y-axis) in ChEBI than either of the earlier studies, and random sets of compounds, over the entire range of ChEBI path lengths (x-axis). Error bars represent one standard deviation over 1000 randomly generated sets.

For interpretation we represent each component by listing the high-valued latent scores z and projection values W. For the latent scores we performed a permutation test to detect the most significantly (q-value < 0.05) activated drugs, while for the projections we inspected the top 30 elements.

## 3 RESULTS

Fig. 2 gives an overview of the types of components discovered by the model. For studying structure-activity relationships, the most important are the components *shared* by one or more chemical view and one or more of the cancer subtypes. The components active in only the expression datasets represent drug responses not captured by the used chemical descriptors, and components only active in the chemical datasets represent biologically irrelevant structural variance. Additionally, components active in only a single dataset may represent dataset-specific noise. We found a total of 11 shared components which will be discussed below. The detailed structure-response relationships discovered from all the shared components are visualized in Supplementary Figure S1 and tabulated in a usable format in Supplementary Table S2.

### 3.1 Validation via Chemical Biology Ontology

We started by quantitatively evaluating how closely related the drugs in the shared components are in terms of known chemical biology relationships, and compared our data with those of two previous studies that investigated drug actions using the CMap database (Iorio *et al.*, 2010; Khan *et al.*, 2012).

The established chemical biology relationships were obtained from the ontology, Chemical Entities of Biological Interest (ChEBI; Degtyarenko *et al.*, 2008), which is the largest such ontology of small compounds. ChEBI links compounds with respect to Chemical structure, Biological roles they are known to play, and their Applications. Examples of classifications are antibiotic, coenzyme and agonist (biological); donor, ligand, inhibitor (chemical), and pesticide, anti-asthmatic (applications). ChEBI was downloaded as a graph containing paths between 328 of our compounds via 611 ontology terms (http://www.ebi.ac.uk/chebi/).

The average similarity (inverse path distance) of drugs within the shared GFA components was consistently higher than the corresponding similarities of Khan *et al.*, (2012), Iorio *et al.*, (2010) and random sets of compounds (Fig. 3). The largest path length (16) in ChEBI linked all drugs, while the smallest (2) linked only the most similar. Interestingly, the difference in GFA and others on small path lengths was higher than that on larger ones, indicating that drugs closely connected in ChEBI were even better found by GFA.

### 3.2 Component Interpretations

We next analyzed the shared components in detail. Each component connects a set of structural drug properties and gene expression changes, forming a hypothesis of a structure activity relationship. A component can be characterized by the set of drugs that activate it the most, and by the set of genes that are expressed differentially when the component is active.

We first compared the findings to the two other studies that have investigated drug actions using the CMap database (Iorio *et al.*, 2010; Khan *et al.*, 2012). Out of the 11 shared GFA components, the majority of the drugs in 7 components were similar to the clusters found by Iorio *et al.* (2010), while 3 components captured structurally driven cell-specific responses they had missed. Compared to the other earlier study (Khan *et al.,* 2012), the majority of the drugs in 6 out of the 11 GFA components matched a corresponding structure-response subcomponent of Khan *et al.,* (2012), again indicating conformance to known results. Our components also revealed several novel drug actions due to cell-type specificity and advanced 3D descriptors that were missed by both of these earlier studies, and are presented below.

Detailed interpretation of all the 11 shared components is presented in Supplementary Table S1. The components were numbered in the order of the amount of variation they captured; the cell line specific components identified by the model were separately ordered with the prefix SP. One component (SP3) captured outlier response of a single drug, and was omitted from further analysis.

The majority of the components captured effects shared among all the three cell lines, while five components had responses which were either cell-line-specific (components SP1, SP2, SP4), dominant in a specific cell line (component 7), or revealed some cell-line-specificity indications for an interesting drug (component 1). The 2D structural features were active in most components, identi-fying similarities in structurally analogous drugs. The pentacle descriptors captured similarities in five components, four of which indicated *novel responses* of drugs that have not been reported before. We discuss these four novel components in detail below. One of these had cell-line-specific effects (SP2), while the remaining cell-line-specific components (SP1 and SP4) are summarized in Table 1.

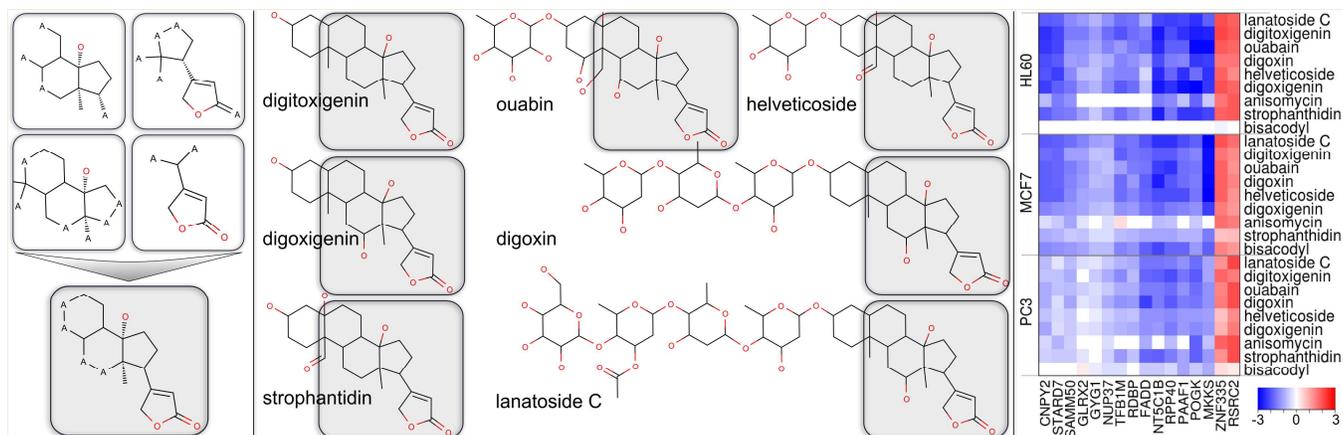

**Fig. 4.** Structure identification in Component 1. **Left**: The top 4 FCFP4 Structural Fragments identified by the model as strongly relating to the response of the drugs (**right**). When combined, these fragments represent the core response triggering structure *steroid backbone* (shaded grey) in all the cardenolides.

**Component 1** was characterized by cardenolides. The top 7 drugs of the component, lanatoside C, digitoxigenin, digoxin, digoxigenin, ouabin, helveticoside and strophantidin belonged to this class. The primary activity of the other drugs anisomycin, lycorine and cicloheximide is protein synthesis inhibition, and bisacodyl is used as a laxative through stimulation of secretion in the colon. Cardenolides act on $Na^+/K^+$ pumps and are known for ion flux alterations. Interestingly, the other compounds of Component 1 also appeared to affect membrane potassium ion flux. Bisacodyl and anisomycin activate $K^+$ flux, lycorine is known to reduce membrane potential (indicative of potassium efflux) and, indicative of affecting $K^+$, emetine needs to be administered with potassium to reduce cardiotoxicity. Interestingly, bisacodyl exhibited the response in MCF7 and PC3 cells only, suggesting that its target may be expressed selectively.

On the structural side, the top 4 FCFP4 fragments collectively represented the correct core 2D response triggering substructure in all the 7 cardenolides, as detailed in Fig. 4. The other two key drugs, bisacodyl and anisomycin, were different from cardiac glycosides in terms of 2D structures, but the Pentacle descriptors indicated potential field similarities on ON, OT and NT probes. These probes referred to existence of common structural pharmacophoric features: hydrogen-bonding and shape-related features. The 3D descriptors may therefore indicate that these drugs do indeed bind the same ion channels as the Cardenolides.

**Component 3** captured protein synthesis inhibition. All drugs in the component are known to inhibit protein synthesis but each in a different way. The only exception, alexidine, is a derivative of clorhexidine, which is used as an antibacterial mouth wash. Interestingly, it has been described to have anti-cancer cell activity through an unknown target (Yip *et al.*, 2006). The model identified pentacle probe fields of D2, DO and DT (shape and lipophilicity related probes) that relate alexidine's protein synthesis inhibition response with the known protein synthesis inhibitors.

**Component 5** was HSP90 inhibition response. The component contains the three very similar drugs geldanamycin, tanespimycin, alvespimycin, and on the 2D structure level very dissimilar 15-delta prostaglandin J2 (PGJ2) and puromycin. Geldanamycin and its two analogs tanespimycin and alvespimycin are heat shock protein (HSP) 90 inhibitors and the latter two have been explored in the clinic as anti-cancer drugs. PGJ2 has also been described as having anti-cancer activity through an unknown mechanism, causing inhibition of several cancer survival signals. Puromycin is reported as an aminonucleoside antibiotic with a primary function of terminating ribosomal protein translation. At the response level, this component appeared to be strongly inducing a heat shock response with many HSP and related genes being up-regulated (see Fig. 5, left). The expression profile very strongly indicated that PGJ2 and puromycin are also inhibiting HSP90. PubChem drug-target data demonstrates that HSP90 targets have been reported as active in geldanamycin and its derivatives, while untested/unspecified for both puromycin and prostaglandin.

On the structural side, the 2D descriptors confirmed that puromycin and prostaglandin are dissimilar to the three geldanamycin analogs. However, the Pentacle descriptors clearly indicated that N2, DN and NT fields shared a very strong pattern across all the five drugs. The patterns were only visible in features of smaller distances of these large molecules, indicating that only a small region of these compounds (polar atoms of all compounds) created the activity, while the rest of the structure is just needed to maintain the shape. This fitted well with the observation that the drugs are overall structurally very dissimilar. At the smaller distances the structure responsible for biological response was characterized by N2: ligands hydrogen bonding capacity, DN: hydrogen bonding and lipopholicity and NT: hydrogen bonding/shape based descriptors. In geldanamycin and prostaglandin this distance (see Fig. 5 where N2 descriptor is plotted) was connected to polar ring-atoms and more precisely corresponding H-bonding positions. These same positions, although in a different conformational arrangement (but with almost identical distance) are critical in the binding of geldanamycin to HSP90. Hence, while the expression data strongly argues for PGJ2 inhibiting HSP90 activity at some level, the structural information does suggest that this effect could be through a direct binding to HSP90 enzymes.

**Table 1.** Shared components having cell-line-specific response. The components (rows) are summarized by their top drugs (column 1), biological response (column 2) and the structural properties (column 3).

| | **Drug Description** | **Biological Interpretation** | **Structural P.** |
| --- | --- | --- | --- |
| **SP1** | Antimetabolite (8-Azaguanine) used for antineoplastic activity and anisomycin a protein synthesis inhibitor. 8-azaguanine has been used in leukemia (Colsky *et al.*, 1955). | Protein synthesis inhibition in HL60 and PC3 cells only. It could be interesting to explore 8-azaguanine as an anti-Prostate Cancer drug. In a recent study Wen *et al.* (2013) also indicated 8-azaguanine for potential therapeutic efficacy in prostate cancer. | 2D Ring structures of 8-azaguanine |
| **SP4** | Anti-Estrogen Drugs | Response visible in MCF7 (estrogen receptor) cell line only. | Pentacle ON/OT fields. |

**Component SP2** is a corticosteroid component, but additionally captured very similar responses by other steroids such as etynodiol as well as the surprisingly different drugs simvastatin and repaglinide. Corticosteroids are anti-inflammatory and typically used against immune responses. Concentrations in the micromolar range were typically used in the CMap data, which is very high, compared to the nanomolar affinities to the corticosteroid receptor primary targets. The response in this component was largely dominant in HL60 cells (Fig. 6) and lightly observed in PC3 while none at all was observed in MCF7, indicating that the relevant target or signal may be selectively expressed in HL60. Both simvastatin (a cholesterol-lowering HMG-CoA inhibitor) and repaglinide (a diabetes drug) are highly dissimilar at the 2D level when compared to the corticosteroids, but both interestingly have been reported to have anti-inflammatory activities, likely due to targets other than the primary target(s). Once again, Pentacle descriptors capture the underlying similarities between these drugs through NT and N2 fields suggesting that the common gene expression patterns induced by the different drugs (corticosteroids, simvastatin and repaglinide) affect the same targets, either primary or off-targets.

## 4 CONCLUSIONS AND DISCUSSION

We extended the drug response analysis paradigm from standard QSAR, of relating drug properties and univariate responses, to finding relationships between specific structural descriptors of drugs with the *genome-wide responses they elicit in multiple cell lines*. The task was formalized as discovering dependencies between multiple datasets, and addressed using the state-of-the-art method Group Factor Analysis (GFA). The approach identified structure-genomic response relationships as underlying components of the data, and can be used as a tool for exploring such relationships from large-scale measurement datasets.

We quantitatively validated our structure-response components over the established chemical-biology relationships of ChEBI and found them to be better than earlier studies (Iorio *et al.,* 2010; Khan *et al.,* 2012) that did not account for separate cell lines and advanced 3D chemical descriptors. Moreover, several drug groups we identified were consistent with earlier studies while several revealed significantly interesting novel findings earlier studies had missed, demonstrating that our approach is viable for explorative multi-set structure-activity analysis. These novel findings were clearly attributed to separating cell line identities and advanced 3D descriptors in our formulation. In a different setting, Yera *et al.,* (2011) found 3D similarity to be more important for off-target identification and this was partially supported by our study as well. However, since the used methods are not identical and since in our approach biological profile (gene expression) was also actively used, this question remains open.

The discovered components revealed interesting new findings of potential importance for revealing novel action mechanisms of drugs. The 2D fingerprints highlighted important core structural groups primarily responsible for activity of similar drugs, such as the identification of the steroid backbone in cardiac glycosides and aromatic ring in HDAC inhibitors. The joint analysis of data from multiple cell lines with advanced 3D Pentacle descriptors allowed us to identify relationships between drugs that were not known earlier. If validated, this suggests an approach that could significantly help in medicinal chemistry and drug design. For example, our data led to the identification of a previously unknown and novel shared mechanism of 15-delta prostaglandin J2 (PGJ2) and heat shock protein (HSP90) inhibitors. Interestingly, PGJ2 and related prostaglandin analogs have repeatedly been described in the literature for having anti-cancer activities, but their mechanism of action has not been clarified before (Fionda *et al.*, 2007; Zimmer *et al.*, 2010; Hegde *et al.*, 2011). Furthermore, our analysis revealed that simvastatin, a cholesterol-lowering drug, has a leukemia-specific anti-inflammatory response very similar to a range of corticosteroids. This appears to be a significant finding in the light that lovastatin, a close structural analog of simvastatin, recently was shown to selectively inhibit leukemic stem cells together with several anti-inflammatory steroids (Hartwell *et al.,* 2013).

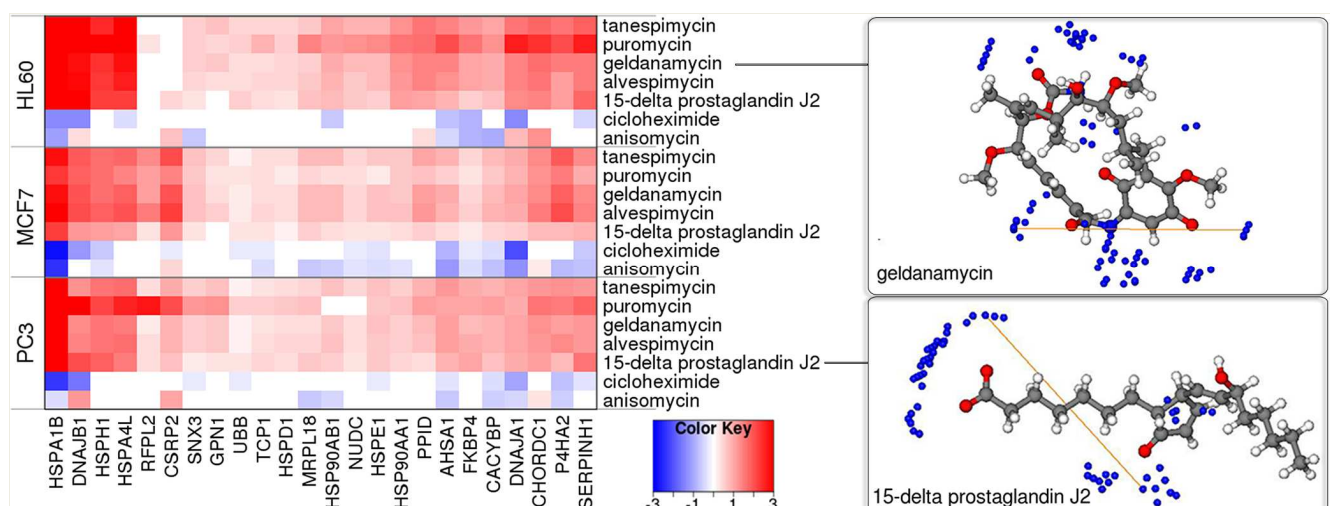

**Fig. 5.** Component 5 identified a novel HSP90 response of prostaglandin. **Left**: Gene expression response of the top 7 drugs in the three cell lines (y-axis), over the top genes (x-axis) of the component, demonstrates HSP genes being strongly up-regulated by the HSP90 inhibitors and by the strikingly very different puromycin and prostaglandin. **Right**: N2 descriptor in geldanamycin and prostaglandin connected to several polar ring-atoms (red and blue). The Pentacle feature (N2 distance range) found by GFA as related with HSP gene expression is represented with the yellow line.

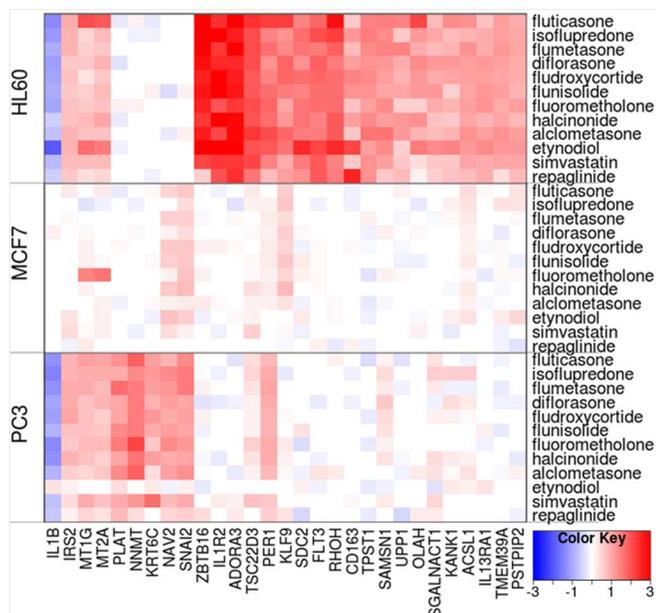

**Fig. 6.** SP2: Corticosteroids showing response specific to HL60 cells, while only minor regulation in PC3 and not at all in MCF7.

Such systematic explorations raise the possibility for targeted interventions and will become a growing trend in the future as more large-scale datasets like the CMap will become available. For drug designers it opens up the opportunity to tailor drug molecules to match a desired gene expression fingerprint. For medicinal chemists, it could help to increase understanding of action mechanisms of existing drugs and revealing potential on-label and off-label applications, including precision medicine.

## ACKNOWLEDGEMENTS

We thank Pekka Tiikkainen for generating the FCFP4 descriptors.

*Funding*: This work was supported by Academy of Finland [140057] and Finnish Centre of Excellence in Computational Inference Research COIN [251170]; the Jane and Aatos Erkko Foundation; and the FICS graduate school.